\documentclass{article}
\usepackage{spconf,amsmath,graphicx}
\usepackage{amsfonts}
\usepackage{siunitx}
\usepackage{adjustbox}
\usepackage{caption}
\usepackage{subcaption}
\usepackage{url}
\usepackage[pscoord]{eso-pic} % To add IEEE's copyright statement

\graphicspath{ {./images/} }
\newcommand{\mysection}[1]{\vspace{-0.2cm}\section{#1}\vspace{-0.2cm}}

\newcommand{\mysubsectionb}[1]{\subsection{#1}\vspace{-0.2cm}}
\newcommand{\mysubsectionc}[1]{\vspace{-0.4cm}\subsection{#1}\vspace{-0.2cm}}

\newcommand{\placetextbox}[4]{% \placetextbox{<offset top>}{<offset left/right>}{<align>}{<stuff>}
  \setbox0=\hbox{#4}% Put <stuff> in a box
  \AddToShipoutPictureFG*{% Add <stuff> to current page foreground
    \if#3r
    \put(\LenToUnit{\paperwidth-#1},\LenToUnit{\paperheight-#2}){\vtop{{\null}\makebox[0pt][r]{\begin{tabular}{r}#4\end{tabular}}}}%
    \else
    \put(\LenToUnit{#1},\LenToUnit{\paperheight-#2}){\vtop{{\null}\makebox[0pt][l]{\begin{tabular}{l}#4\end{tabular}}}}%
    \fi
  }%
}%
\placetextbox{0.2cm}{0.2cm}{l}{\textcopyright 2024 IEEE. Personal use of this material is permitted. Permission from IEEE must be obtained for all other uses, in any current or future media, \\ including reprinting/republishing this material for advertising or promotional purposes, creating new collective works, for resale or redistribution \\ to servers or lists, or reuse of any copyrighted component of this work in other works.}

\title{HYB-VITON: A HYBRID APPROACH TO VIRTUAL TRY-ON COMBINING EXPLICIT AND IMPLICIT WARPING}
\name{Kosuke Takemoto, Koshinaka Takafumi}
\address{The Graduate School of Data Science, Yokohama City University, Japan}

\begin{document}
%\ninept
%
\maketitle

\begin{abstract}
Virtual try-on systems have significant potential in e-commerce,
allowing customers to visualize garments on themselves.
Existing image-based methods fall into two categories:
those that directly warp garment-images onto person-images (explicit warping),
and those using cross-attention to reconstruct given garments (implicit warping).
Explicit warping preserves garment details but often produces unrealistic output,
while implicit warping achieves natural reconstruction but struggles with fine details.
We propose HYB-VITON, a novel approach that combines the advantages of each method and
includes both a preprocessing pipeline for warped garments and a novel training option.
These components allow us to utilize beneficial regions of explicitly warped garments
while leveraging the natural reconstruction of implicit warping.
A series of experiments demonstrates
that HYB-VITON preserves garment details more faithfully than recent diffusion-based methods,
while producing more realistic results than a state-of-the-art explicit warping method.
\end{abstract}
\mysection{INTRODUCTION}
\label{sec:intro}

The rise of e-commerce in the fashion industry has created a strong demand
for the development of virtual try-on systems that enhance the online shopping experience by allowing
customers to visualize clothing items being worn on themselves, potentially increasing satisfaction and reducing return rates.
Typical image-based virtual try-on methods use a person-image
and a garment-image as input in the generation of an output image of
the person wearing the garment. Two main paradigms exist in this field. One is based on explicit warping,
typically using Generative Adversarial Networks (GANs),
while the other uses implicit warping performed through cross-attention among diffusion models.

Explicit warping models first explicitly transform the garment-image to match the person’s pose,
using such techniques as Thin-Plate Spline (TPS) or flow estimation \cite{clothflow, gpvton, kgi}.
The warped garment is then fed into a generative model to produce the final output image.
While these approaches can preserve garment details effectively when warping is successful,
they often struggle to produce realistic results due to the limited regions of the garment that are shown in a garment-image.
Moreover, there are some known artifacts, including the lack of natural wrinkles that occur when wearing clothes,
squeezing or stretching effects near garment boundaries, and backing fabrics around the neck.
\begin{figure}[t]
  \label{fig:top_image}
    \captionsetup{belowskip=-0.5cm}
  \includegraphics[width=\linewidth]{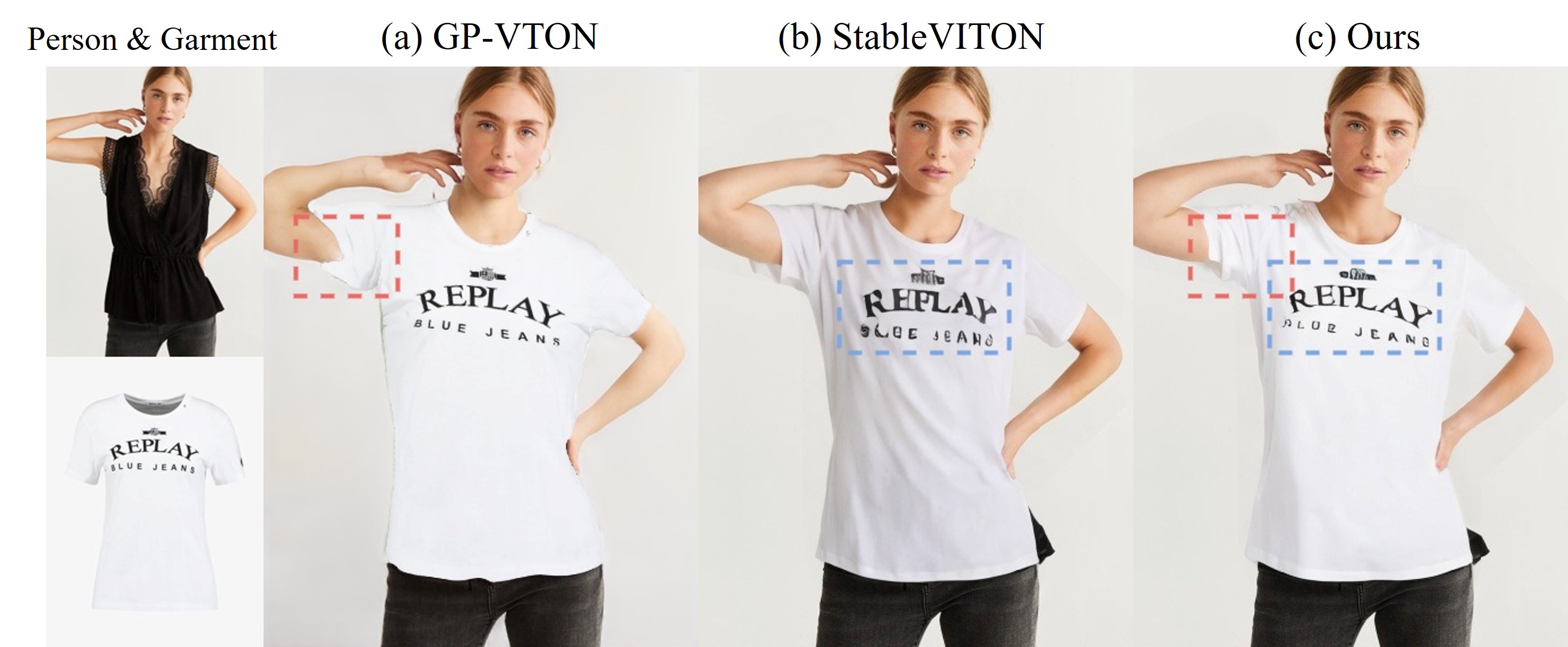}
  \caption{Our hybrid approach (c) better preserves fine details of garments, such as text and logos, \
  than does current implicit warping (b), \
  while eliminating known artifacts of explicit warping (a).}
\end{figure}

In response to these challenges,
recent methods have employed implicit warping techniques using diffusion models \cite{tryondiffusion, stableviton}.
These models gradually transform random noise into realistic images, guided by the person- and garment-images.
They reconstruct a garment through cross-attention, which generally results in a more natural output.
However, these approaches often struggle to preserve such fine details as printed illustration, text, and logos,
which are crucial for garment fidelity, as they often represent main features and/or the brands of garments.
This trade-off between natural reconstruction and detail preservation highlights the need
for a new approach that utilizes the advantages of both explicit and implicit warping methods.

Some recent methods use a pre-trained explicit warping model
in conjunction with cross-attention to improve the quality of output images \cite{ladi-vton, cat-dm}.
These methods, however, incorporate explicitly warped garments as a prior
for a diffusion model by adding them to input and/or using them to make starting points for a denoising process.
Since there are discrepancies between the ground-truth garment in the target image
and a pre-warped garment produced by such an external model,
the diffusion model is still responsible for moving garment pixels to their correct locations.
As a result, these methods do not fully leverage the benefits of explicit warping for detail preservation.

To fully leverage the pattern placement capabilities of explicit warping,
we address two key challenges: first, identifying the beneficial regions of warped garments in light of known artifacts;
and second, developing a method by which our network can utilize warped garments beyond their current role as priors,
thus enhancing garment fidelity.
In response to these challenges, we propose HYB-VITON
\footnote{The source code will be available at \url{https://github.com/takesukeDS/HYB-VITON}.},
which includes a preprocessing pipeline for warped garments
and a novel training option for virtual try-on networks.
Experimental results demonstrate
that our approach not only preserves garment features more faithfully than recent diffusion-based methods
but also produces more realistic results than a state-of-the-art explicit warping method.
These findings contribute to the field by both encouraging further advancements in explicit warping methods
and establishing a benchmark for detail preservation in future implicit warping development.

\mysection{PROPOSED METHOD}
\label{sec:proposed}

We adopt a conventional training setting,
utilizing paired data consisting of a flattened garment-image $C \in \mathbb{R}^{3 \times H \times W}$
and an image of a person $I \in \mathbb{R}^{3 \times H \times W}$ wearing that garment.
Our method employs a diffusion model trained to inpaint the missing region of a garment-agnostic person-image $I_\text{a}$,
which is derived from an original person-image by removing the garment and the area around it.
The mask $M_\text{a} \in \{0, 1\}^{1 \times H \times W}$ represents the region to be preserved.
For actual virtual try-on scenarios, we employ an unpaired setting
in which the network is provided with a different garment $C'$ from the one the person is wearing in the input image $I$.

Explicit warping methods utilize the garment mask $M \in \{0, 1\}^{1 \times H \times W}$,
which denotes the garment region in the original garment-image $C$.
These methods produce a warped garment-image $C^\text{w}_\text{EX}$ and a warped mask $M^\text{w}_\text{EX}$.
In addition to such conventional input, we introduce $C^\text{w}_\text{GT}$,
the garment region of the ground truth person-image $I$.
By using $C^\text{w}_\text{GT}$ rather than $C^\text{w}_\text{EX}$ during training,
our model is encouraged to utilize the pattern placement of the given warped garment more effectively.

Our network is based on Stable Diffusion (SD) \cite{ldm},
an open-source text-to-image model
which trains a diffusion model on the latent space of a pre-trained Variational Autoencoder (VAE) \cite{vae}.
In SD, all input images are encoded into the latent space
using the encoder $\mathcal{E}(\cdot)$ before feeding them into the network.
Additionally, input masks are resized to match the spatial dimensions of the latent.

\mysubsectionc{Preprocessing of warped garments}
\label{ssec:proposed-hybrid}
To deal with the noticeable artifacts often present in explicitly warped garments,
we implement a crucial step to extract the usable region, thereby safely using the warped garments for garment fidelity.
We apply the following three-step process to the warped garment $C^\text{w}_\text{EX}$ at inference,
and to the ground truth garment $C^\text{w}_\text{GT}$ at training,
producing preprocessed garment $C^\text{w}$ and its mask $M^\text{w}$:
\vspace{-1.35mm}
\begin{enumerate}
  \itemsep0em
  \item \textbf{Torso extraction}:
  We utilize the semantic segmentation $S$ produced by DensePose \cite{densepose} to extract the torso region of the garment.
  This step is particularly important for long-sleeved tops,
  for which significant misalignments can occur
  between the sleeve regions in the warped garment $C^\text{w}_\text{EX}$ and those in the person-image,
  especially when the person's arms are bent.
  \item \textbf{Region erosion}:
  We erode the torso region of the garment to eliminate regions
  affected by squeezing or stretching near garment boundaries,
  as well as regions that should be occluded by the person's neck in the output image.
  This erosion is performed by applying a minimum filter
  to the masks $M^\text{w}_\text{EX}$ and $M^\text{w}_\text{GT}$.
  \item \textbf{Edge-preserving filtering}:
  We apply a bilateral filter \cite{bilateral}, an edge-preserving filter, to the remaining region.
  This step eliminates wrinkles while maintaining the pattern placement of the garment.

\end{enumerate}
\vspace{-1.35mm}
Regarding our approach of using $C^\text{w}_\text{GT}$ rather than $C^\text{w}_\text{EX}$ during training,
it is important to note that the wrinkles in $C^\text{w}_\text{GT}$ can be problematic.
This is because the warped garment $C^\text{w}_\text{EX}$ does not possess wrinkles
that would be created when actually wearing the garment.
Therefore, we apply the bilateral filter in Step 3 to encourage the network to infer natural wrinkles
rather than simply replicating the ground truth wrinkles in $C^\text{w}_\text{GT}$.
The kernel for a pixel located at $(i,j)$ in RGB channel $c$ has the following weights:
\begin{eqnarray}
    A \exp \left(-\frac{(i - k)^2 + (j - l)^2}{2\sigma_\text{d}^2} - \frac{(I_c(i, j) - I_c(k, l))^2}{2\sigma_\text{r}^2}\right),
    \nonumber
\end{eqnarray}
where $(k, l)$ is the location of its neighboring pixel,
$A$ is the normalizing constant, $I_c(i, j)$ is the pixel value at $(i, j)$ in channel $c$.
The parameters $\sigma_\text{d}$ and $\sigma_\text{r}$ determine the trade-off between preserving garment features and removing wrinkles.

We use different values for $\sigma_\text{r}$ during inference than were used in training,
as the warped garment $C^\text{w}_\text{EX}$ has fewer wrinkles than the ground truth garment $C^\text{w}_\text{GT}$.
We explored these values using our validation set and found that a smaller $\sigma_\text{r}$ could be used at inference.
\mysubsectionc{Hybrid of explicit and implicit warping}
\label{ssec:proposed-hybrid}
\begin{figure}[t]
  \centering
  \includegraphics[width=\linewidth]{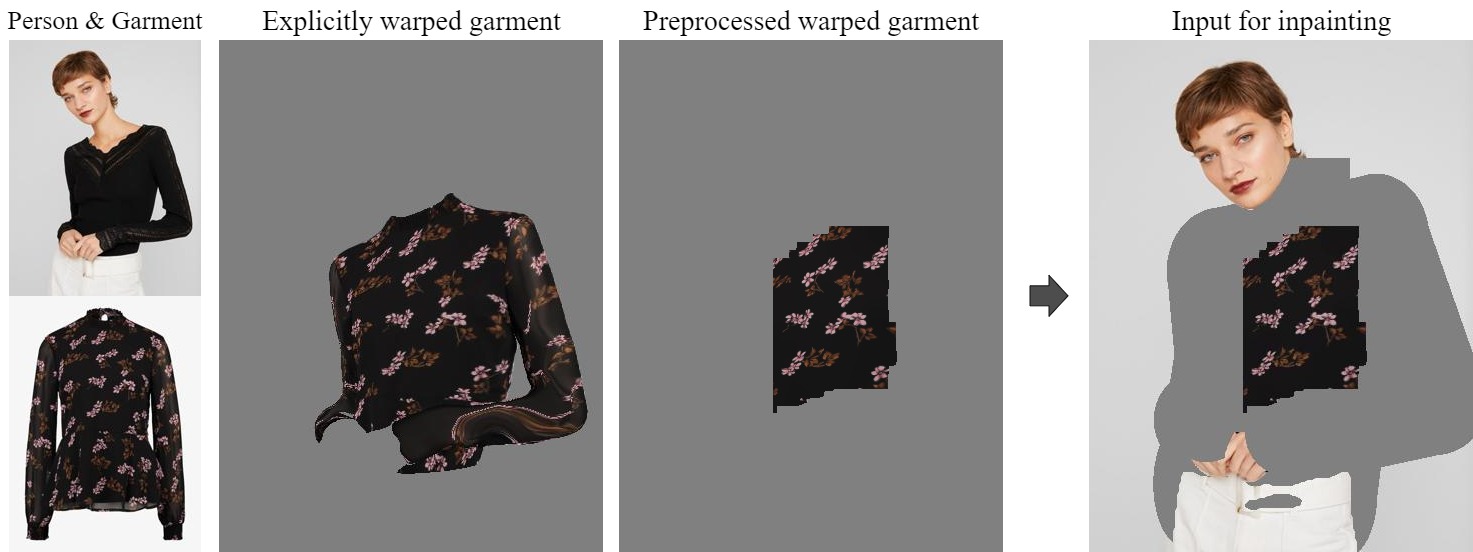}
  \caption{Our method first preprocesses a warped garment to form an input for inpainting.}
  \label{fig:process}
\end{figure}
\begin{figure}[t]
    \centering
    \captionsetup{belowskip=-0.3cm}
    \includegraphics[width=0.9\linewidth]{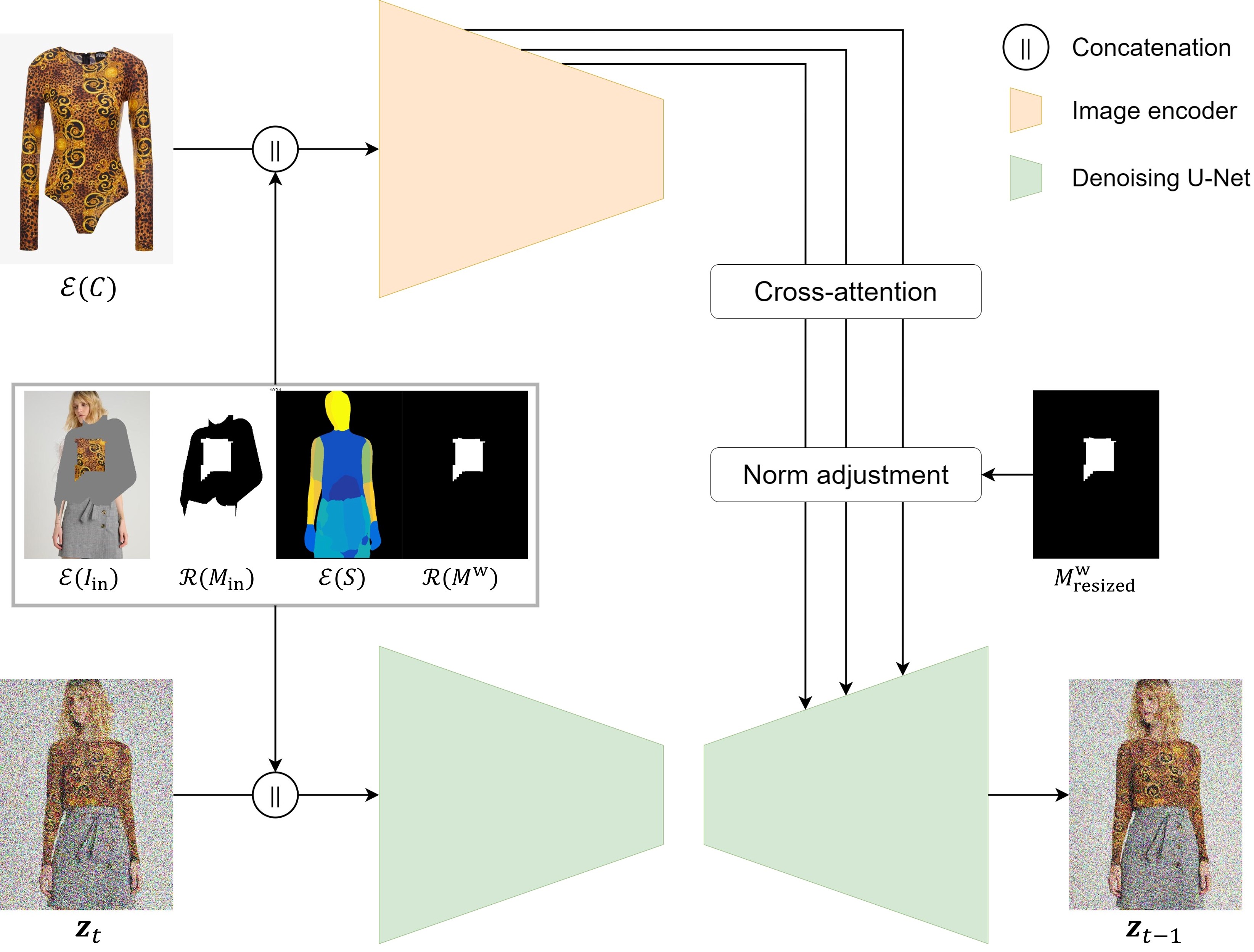}
    \caption{A denoising step of HYB-VITON.}
    \label{fig:unet}
\end{figure}
As shown in Fig.\,\ref{fig:process},
we incorporate the preprocessed garment $C^\text{w}$ into the network by adding it to $I_\text{a}$,
forming an input tensor $I_\text{in} = I_\text{a} + (1 - M_\text{a}) \odot C^\text{w}$
and an input mask $M_\text{in} = M_\text{a} + (1 - M_\text{a}) \odot M^\text{w}$,
where $\odot$ denotes an element-wise multiplication.
% as if the warped garment is part of the region to be preserved.
Additionally, we include the mask $M^\text{w}$ in the input for the denoising U-Net \cite{unet},
enabling the network to distinguish the region occupied by the preprocessed garment in $I_\text{in}$.
The input to the denoising U-Net in Fig.\,\ref{fig:unet}
at diffusion time step $t$ consists of
the noisy latent $\boldsymbol{z}_t$, $\mathcal{E}(I_\text{in})$, $\mathcal{R}(M_\text{in})$, $\mathcal{R}(M^\text{w})$,
and the semantic segmentation of the person's body $\mathcal{E}(S)$,
where $\mathcal{R}(\cdot)$ denotes the resizing operation.

Concurrently, the original garment-image $C$ is fed into the image encoder along with other person representations.
The internal representations of the garment are then passed to the denoising U-Net via cross-attention.
However, since the spatially aligned garment is already present in $I_\text{in}$,
this implicit warping could conflict with the explicit warping.
To address this issue, we adjust the norm of the cross-attention output in the warped garment region
with a new learnable scalar $\alpha^l$ for each layer $l$ with the cross-attention.
The implicit warping is adjusted using $\alpha^l$ and the warped mask $M^\text{w}$ as follows:
\begin{eqnarray}
    (1 - \alpha^l M^\text{w}_\text{resized}) \odot \text{Attn}^l_{\text{2D}},
    \nonumber
\end{eqnarray}
where $M^\text{w}_\text{resized} \in \{0,1\}^{1 \times h^l \times w^l}$ is a resized version of $M^\text{w}$
to match the spatial dimensions of the feature maps $(h^l,\,w^l)$ in the layer,
and $\text{Attn}^l_\text{2D} \in \mathbb{R}^{c^l \times h^l \times w^l}$ is the output of the layer having the cross-attention.

\mysection{EXPERIMENTS}
\label{sec:experiments}
\begin{figure*}[t]
    \label{fig:existing_methods_comp}
    \centering
    \captionsetup{width=0.9\textwidth, belowskip=0cm}
    \begin{subfigure}[b]{0.575\textwidth}
        \captionsetup{width=\textwidth}
        \includegraphics[width=\textwidth, keepaspectratio]{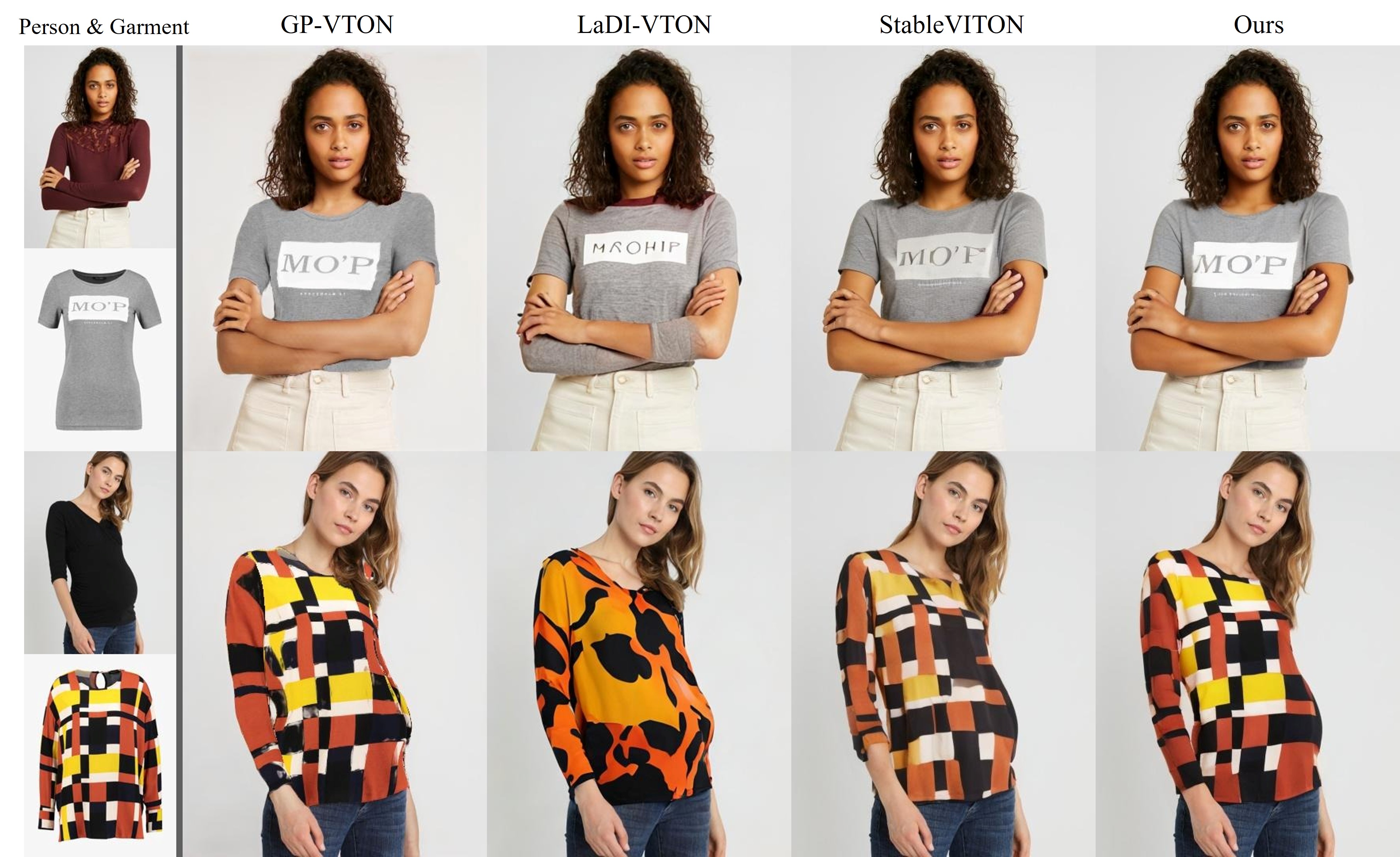}
        \caption{Comparisons in unpaired setting.}
        \label{fig:unpaired}
    \end{subfigure}
    \hspace{\fill}
    \begin{subfigure}[b]{0.375\textwidth}
        \captionsetup{width=\textwidth, belowskip=0cm}
        \includegraphics[width=\textwidth, keepaspectratio]{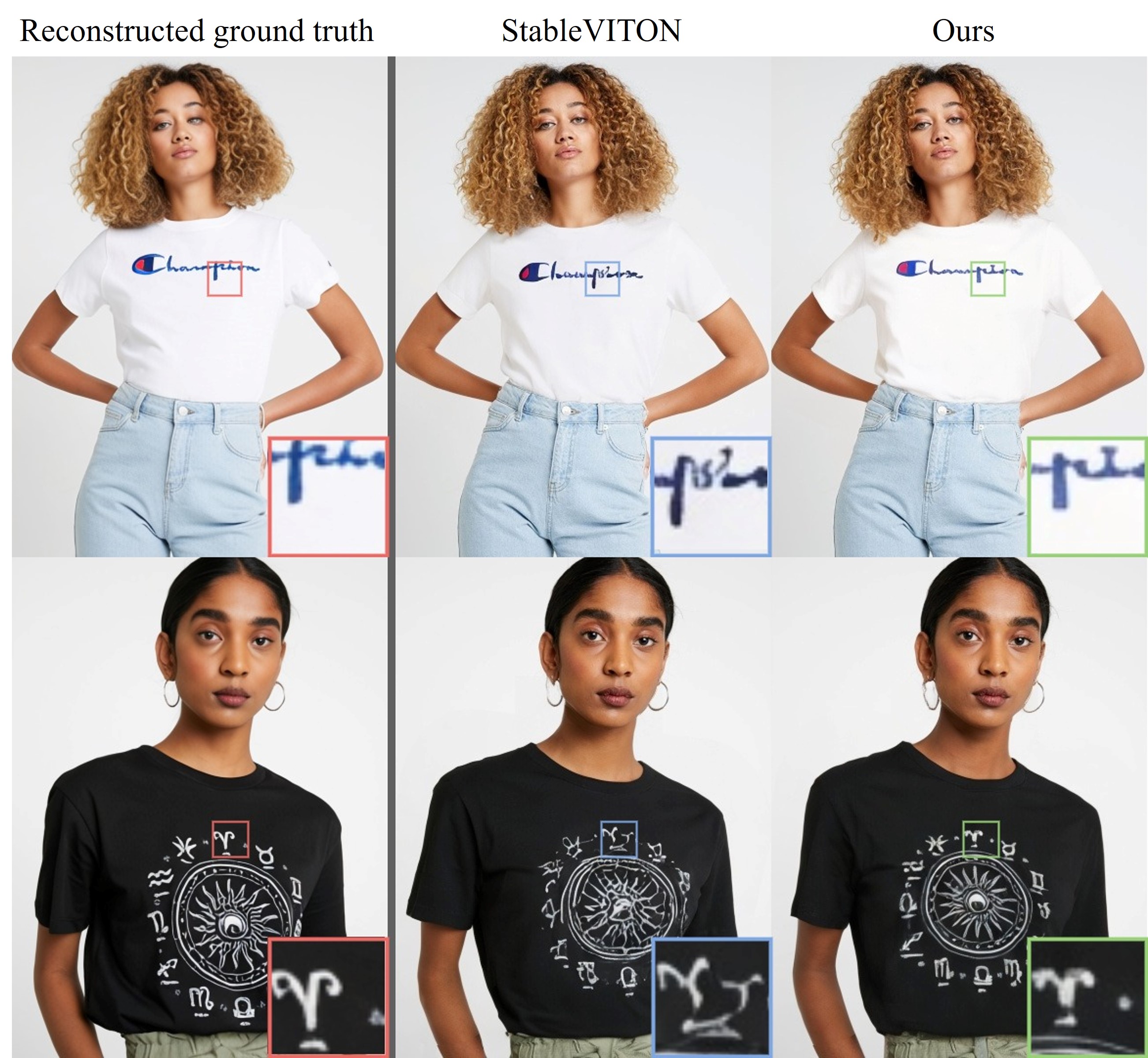}
        \caption{Fine detail comparisons in paired setting.}
        \label{fig:paired}
    \end{subfigure}
    \caption{Qualitative comparisons. Best viewed when zoomed in on.}
    \label{fig:existing_methods_comp_unpaired}
\end{figure*}
\mysubsectionb{Experiment setup}
\label{ssec:experiments-setup}
We conducted experiments on the VITON-HD dataset \cite{vitonhd}, a popular virtual try-on dataset featuring high-resolution images.
While our method can handle a range of resolutions,
we used images resized to $512 \times 384$ throughout our experiments,
consistent with many previous research projects.
The VITON-HD dataset contains \num[group-separator={,}]{11647} training pairs and \num[group-separator={,}]{2032} test pairs.
For validation purposes, we randomly selected \num[group-separator={,}]{1000} pairs from the training set.
Our method builds upon the architecture of StableVITON \cite{stableviton},
a state-of-the-art implicit warping method based on Stable Diffusion (SD).
We extended the input channels and initialized the network with pre-trained weights.
We fine-tuned the input layers of both the image encoder and the denoising U-Net,
as well as all decoder blocks of the denoising U-Net, while keeping other parameters frozen.
The network was trained using the AdamW optimizer with a learning rate of $10^{-4}$ for 100 epochs, using a batch size of 32.

We used explicitly warped garments from GP-VTON \cite{gpvton}, a state-of-the-art explicit warping method.
In our preprocessing pipeline, we applied two filters: a minimum filter and a bilateral filter.
For the minimum filter, we set the kernel size to $21 \times 21$.
The bilateral filter used a kernel size of $23 \times 23$, with $\sigma_\text{d} = 5$ and $\sigma_\text{r} = 0.06$ during training;
at inference time, we adjusted $\sigma_\text{r}$ to 0.01. Additionally, we set the initial value of $\alpha^l$ to 0.5.

Virtual try-on can be approached as an inpainting task.
In this context, a virtual try-on model can initiate the denoising process
with the perturbed version of a garment-agnostic person-image $I_\text{a}$, similar to SDEdit \cite{sdedit}.
Moreover, our hybrid approach incorporated the warped garment into the network,
allowing us to start the denoising process with the perturbed version of $I_\text{in}$,
which included the preprocessed warped garment $C^\text{w}$.
We employed a pseudo linear multi-step (PLMS) \cite{plms} sampler with 50 steps, matching the setting used in StableVITON.

We compared our method with GP-VTON, LaDI-VTON \cite{ladi-vton}, and StableVITON.
Among these, LaDI-VTON adopts an SD-based hybrid approach with a pre-trained image encoder.

For quantitative evaluation,
we employed standard metrics widely used in the virtual try-on field.
For the paired setting,
in which we could directly compare the generated image with the ground truth image,
we used the Structural Similarity Index (SSIM) \cite{ssim} and
Learned Perceptual Image Patch Similarity (LPIPS) \cite{lpips}.
For the unpaired setting, we used the Fr\'echet Inception Distance (FID) \cite{fid} and
the Kernel Inception Distance (KID) \cite{kid} to compare the distribution of the real images
with that of the generated images.
\mysubsectionc{Qualitative results}
\label{ssec:experiments-qualitative}
Fig.\,\ref{fig:unpaired} shows comparisons of our method against state-of-the-art approaches in the unpaired setting.
In the upper row, GP-VTON, using an explicitly warped garment without preprocessing, preserves the logo well,
but their result lacks natural wrinkles and shades.
StableVITON produces a natural image, but fails to preserve fine details.
In contrast to this, our method successfully preserves the logo while producing natural wrinkles.
In the lower row, GP-VTON exhibits squeezing/stretching effects near the garment boundaries,
causing obvious artifacts.
Our method, however, successfully removes these artifacts while preserving garment features.
In both cases, LaDI-VTON struggles to preserve the identity of the garment,
indicating that a pre-trained image encoder is not sufficient to maintain garment features.
Moreover, it does not utilize the warped garment effectively for pattern placement.

Fig.\,\ref{fig:paired} illustrates the detail preservation capabilities of our method in the paired setting. % check space
We compare here HYB-VITON with StableVITON and the person-image reconstructed by the pre-trained VAE.
While garment reconstruction solely using cross-attention yields unstable results,
our method successfully preserves the fine details of the garment with only slight modifications.
\mysubsectionc{Quantitative results}
\label{ssec:experiments-quantitative}
\begin{table}[t]
    \captionsetup{belowskip=-0.2cm, aboveskip=0.2cm}
    \centering
    \caption{Quantitative comparison of virtual try-on methods.}
  \begin{adjustbox}{width=\linewidth,center}
    \begin{tabular}{ c c c c c } \hline
Method &  SSIM↑ & LPIPS↓ & $\text{FID}_{\text{u}}$↓ & $\text{KID}_{\text{u}}$↓ \\ \hline
GP-VTON \cite{gpvton} & 0.873 & 0.0841 &  9.67 & 1.496 \\
LaDI-VTON \cite{ladi-vton} & 0.856 & 0.0917 & 9.363 & 1.609 \\
StableVITON \cite{stableviton} & \textbf{0.874} & 0.0759 & 9.128 & 1.270 \\ \hline
Ours & 0.873 & \textbf{0.0751} & \textbf{9.023} & \textbf{1.045}  \\ \hline
    \end{tabular}
  \end{adjustbox}
    \label{tab:quantitative}
    \vspace{-0.4cm}
\end{table}
As shown in Table \ref{tab:quantitative}, HYB-VITON outperforms all baselines in the unpaired setting,
achieving the best FID and KID scores.
In the paired setting, our method slightly improves the LPIPS score over that of StableVITON,
indicating better perceptual similarity to ground truth images.
However, the SSIM score remains comparable to GP-VTON and slightly lower than StableVITON.
This suggests that explicitly warped garments do not provide significant additional information
for an implicit warping model to reconstruct the luminance and/or the contrast of a person-image.
\mysubsectionc{Ablation study}
\label{ssec:experiments-ablation}
We conducted an ablation study to confirm the effectiveness of the norm adjustment.
We compared our full HYB-VITON model against two variants:
1) a model without norm adjustment, and
2) a model with norm adjustment but with the scalar $\alpha^l$ fixed to 1,
disabling implicit warping for the warped garment region.
\begin{table}[t]
    \captionsetup{belowskip=-0.2cm, aboveskip=0.2cm}
    \centering
    \caption{Ablation study of the norm adjustment.}
  \begin{adjustbox}{width=\linewidth,center}
    \begin{tabular}{ c c c c c } \hline
Method &  SSIM↑ & LPIPS↓ & $\text{FID}_{\text{u}}$↓ & $\text{KID}_{\text{u}}$↓ \\ \hline
w/o Norm adjustment & 0.872 & 0.0761 & 9.255 & 1.313 \\
w/o Implicit warping & 0.873 & 0.0767 & 9.182 & 1.163 \\ \hline
HYB-VITON (full) & 0.873 & 0.0751 & 9.023 & 1.045 \\ \hline
    \end{tabular}
  \end{adjustbox}
  \label{tab:ablation}
  \vspace{-0.4cm}
\end{table}

As shown in Table \ref{tab:ablation},
the model without norm adjustment exhibits a decrease in performance across all metrics,
indicating the conflict between explicit and implicit warping.
Disabling implicit warping for the warped garment region avoids this conflict.
Since the filtering in our preprocessing pipeline
causes the warped garment to lose some information (e.g., the type of material),
however, we need to complement the garment region with implicit warping.

% To start a new column (but not a new page) and help balance the last-page
% column length use \vfill\pagebreak.
% -------------------------------------------------------------------------
%\vfill
%\pagebreak
\vspace{-1mm}
\mysection{SUMMARY}
\label{sec:conclusion}
We have proposed here HYB-VITON, a novel hybrid approach to image-based virtual try-on
that effectively combines explicit and implicit warping techniques.
Our proposed preprocessing pipeline enables the network to utilize the advantages of explicit warping for detail preservation,
while maintaining overall naturalness in the final output image.
Our experiments demonstrate that HYB-VITON outperforms recent diffusion-based methods
in terms of detail preservation from both qualitative and quantitative perspectives.
%, despite using only limited regions of the warped garments.
In future research, improving explicit warping techniques to expand the usable regions of warped garments
could further enhance the performance of hybrid approaches like HYB-VITON.

\noindent
\textbf{Limitations.}
Our model exhibits infrequent failure cases,
stemming from poorly transformed garments and limited generative ability for missing garment regions.

\noindent
\textbf{Acknowledgments.} This work was partially supported by JSPS KAKEN Grant Numbers 21K11967 and 24K15012.

\vfill\pagebreak

% References should be produced using the bibtex program from suitable
% BiBTeX files (here: strings, refs, manuals). The IEEEbib.bst bibliography
% style file from IEEE produces unsorted bibliography list.
% -------------------------------------------------------------------------
\bibliographystyle{IEEEbib}
\bibliography{refs}

\begin{thebibliography}{10}

\bibitem{clothflow}
Xintong Han, Weilin Huang, Xiaojun Hu, and Matthew Scott,
\newblock ``{ClothFlow}: A flow-based model for clothed person generation,''
\newblock in {\em 2019 IEEE/CVF International Conference on Computer Vision
  (ICCV)}, 2019.

\bibitem{gpvton}
Xie Zhenyu, Huang Zaiyu, Dong Xin, Zhao Fuwei, Dong Haoye, Zhang Xijin, Zhu
  Feida, and Liang Xiaodan,
\newblock ``{GP-VTON}: Towards general purpose virtual try-on via collaborative
  local-flow global-parsing learning,''
\newblock in {\em Proceedings of the IEEE/CVF Conference on Computer Vision and
  Pattern Recognition (CVPR)}, June 2023.

\bibitem{kgi}
Zhi Li, Pengfei Wei, Xiang Yin, Zejun Ma, and Alex~C. Kot,
\newblock ``Virtual try-on with pose-garment keypoints guided inpainting,''
\newblock in {\em Proceedings of the IEEE/CVF International Conference on
  Computer Vision (ICCV)}, October 2023.

\bibitem{tryondiffusion}
Luyang Zhu, Dawei Yang, Tyler Zhu, Fitsum Reda, William Chan, Chitwan Saharia,
  Mohammad Norouzi, and Ira Kemelmacher-Shlizerman,
\newblock ``{TryOnDiffusion}: A tale of two unets,''
\newblock in {\em Proceedings of the IEEE/CVF Conference on Computer Vision and
  Pattern Recognition (CVPR)}, June 2023.

\bibitem{stableviton}
Jeongho Kim, Guojung Gu, Minho Park, Sunghyun Park, and Jaegul Choo,
\newblock ``{StableVITON}: Learning semantic correspondence with latent
  diffusion model for virtual try-on,''
\newblock in {\em Proceedings of the IEEE/CVF Conference on Computer Vision and
  Pattern Recognition (CVPR)}, June 2024.

\bibitem{ladi-vton}
Davide Morelli, Alberto Baldrati, Giuseppe Cartella, Marcella Cornia, Marco
  Bertini, and Rita Cucchiara,
\newblock ``{LaDI-VTON}: Latent diffusion textual-inversion enhanced virtual
  try-on,''
\newblock in {\em Proceedings of the 31st ACM International Conference on
  Multimedia}, 2023, MM '23.

\bibitem{cat-dm}
Jianhao Zeng, Dan Song, Weizhi Nie, Hongshuo Tian, Tongtong Wang, and An-An
  Liu,
\newblock ``{CAT-DM}: Controllable accelerated virtual try-on with diffusion
  model,''
\newblock in {\em Proceedings of the IEEE/CVF Conference on Computer Vision and
  Pattern Recognition (CVPR)}, June 2024.

\bibitem{ldm}
Robin Rombach, Andreas Blattmann, Dominik Lorenz, Patrick Esser, and Bj\"orn
  Ommer,
\newblock ``High-resolution image synthesis with latent diffusion models,''
\newblock in {\em Proceedings of the IEEE/CVF Conference on Computer Vision and
  Pattern Recognition (CVPR)}, June 2022.

\bibitem{vae}
Diederik~P. Kingma and Max Welling,
\newblock ``{Auto-Encoding Variational Bayes},''
\newblock in {\em 2nd International Conference on Learning Representations,
  {ICLR} 2014, Banff, AB, Canada, April 14-16, 2014, Conference Track
  Proceedings}, 2014.

\bibitem{densepose}
Rıza~Alp Güler, Natalia Neverova, and Iasonas Kokkinos,
\newblock ``{DensePose}: Dense human pose estimation in the wild,''
\newblock in {\em Proceedings of the IEEE Conference on Computer Vision and
  Pattern Recognition (CVPR)}, June 2018.

\bibitem{bilateral}
C.~Tomasi and R.~Manduchi,
\newblock ``Bilateral filtering for gray and color images,''
\newblock in {\em Sixth International Conference on Computer Vision (IEEE Cat.
  No.98CH36271)}, 1998.

\bibitem{unet}
Olaf Ronneberger, Philipp Fischer, and Thomas Brox,
\newblock ``{U-Net}: Convolutional networks for biomedical image
  segmentation,''
\newblock in {\em Medical Image Computing and Computer-Assisted Intervention --
  MICCAI 2015}, 2015.

\bibitem{vitonhd}
Seunghwan Choi, Sunghyun Park, Minsoo Lee, and Jaegul Choo,
\newblock ``{VITON-HD}: High-resolution virtual try-on via misalignment-aware
  normalization,''
\newblock in {\em Proc. of the IEEE conference on computer vision and pattern
  recognition (CVPR)}, 2021.

\bibitem{sdedit}
Chenlin Meng, Yutong He, Yang Song, Jiaming Song, Jiajun Wu, Jun-Yan Zhu, and
  Stefano Ermon,
\newblock ``{SDE}dit: Guided image synthesis and editing with stochastic
  differential equations,''
\newblock in {\em International Conference on Learning Representations}, 2022.

\bibitem{plms}
Luping Liu, Yi~Ren, Zhijie Lin, and Zhou Zhao,
\newblock ``Pseudo numerical methods for diffusion models on manifolds,''
\newblock in {\em International Conference on Learning Representations}, 2022.

\bibitem{ssim}
Zhou Wang, A.C. Bovik, H.R. Sheikh, and E.P. Simoncelli,
\newblock ``Image quality assessment: from error visibility to structural
  similarity,''
\newblock {\em IEEE Transactions on Image Processing}, 2004.

\bibitem{lpips}
Richard Zhang, Phillip Isola, Alexei~A. Efros, Eli Shechtman, and Oliver Wang,
\newblock ``The unreasonable effectiveness of deep features as a perceptual
  metric,''
\newblock in {\em Proceedings of the IEEE Conference on Computer Vision and
  Pattern Recognition (CVPR)}, June 2018.

\bibitem{fid}
Martin Heusel, Hubert Ramsauer, Thomas Unterthiner, Bernhard Nessler, and Sepp
  Hochreiter,
\newblock ``{GAN}s trained by a two time-scale update rule converge to a local
  nash equilibrium,''
\newblock in {\em Advances in Neural Information Processing Systems}, 2017.

\bibitem{kid}
Mikołaj Bińkowski, Dougal~J. Sutherland, Michael Arbel, and Arthur Gretton,
\newblock ``Demystifying {MMD} {GAN}s,''
\newblock in {\em International Conference on Learning Representations}, 2018.

\end{thebibliography}

\end{document}